\documentclass{article}
\usepackage{ijcai24}

\usepackage{times}
\usepackage{soul}
\usepackage{url}
\usepackage[hidelinks]{hyperref}
\usepackage[utf8]{inputenc}
\usepackage[small]{caption}
\usepackage{graphicx}
\usepackage{amsmath}
\usepackage{amsthm}
\usepackage{booktabs}
\usepackage{algorithm}
\usepackage{algorithmic}
\usepackage[switch]{lineno}

\usepackage{amssymb}
\usepackage{multirow}
\usepackage{subfigure}
\usepackage{pifont}
\usepackage[T1]{fontenc}
\usepackage{newfloat}
\usepackage{listings}

\usepackage[table,xcdraw]{xcolor}

\title{Controlling Large Language Model-based Agents for Large-Scale Decision-Making: An Actor-Critic Approach}

\author{
Bin Zhang$^{1,2}$
\and
Hangyu Mao$^{3,*}$\and
Jingqing Ruan$^{1,2}$\and
Ying Wen$^4$\and
Yang Li$^5$\and
Shao Zhang$^4$\and\\  
Zhiwei Xu$^{1,2}$\and
Dapeng Li$^{1,2}$\and
Ziyue Li$^3$\and
Rui Zhao$^3$\and
Lijuan Li$^{1,2,}$\thanks{Corresponding author: \{hy.mao@pku.edu.cn, maohangyu@sensetime.com\}, lijuan.li@ia.ac.cn}\And
Guoliang Fan$^{1,2}$\\
\affiliations
$^1$Institute of Automation,Chinese Academy of Sciences\\
$^2$School of Artificial Intelligence, University of Chinese Academy of Sciences\\
$^3$SenseTime Research\\
$^4$Shanghai Jiao Tong University\\
$^5$The University of Manchester\\
}

\begin{document}

\maketitle

\begin{abstract}
The remarkable progress in Large Language Models (LLMs) opens up new avenues for addressing planning and decision-making problems in Multi-Agent Systems (MAS). However, as the number of agents increases, the issues of hallucination in LLMs and coordination in MAS have become increasingly prominent.
Additionally, the efficient utilization of tokens emerges as a critical consideration when employing LLMs to facilitate the interactions among a substantial number of agents. 
In this paper, we develop a modular framework called LLaMAC to mitigate these challenges. LLaMAC implements a value distribution encoding similar to that found in the human brain, utilizing internal and external feedback mechanisms to facilitate collaboration and iterative reasoning among its modules.
Through evaluations involving system resource allocation and robot grid transportation, we demonstrate the considerable advantages afforded by our proposed approach.
\end{abstract}

\section{Introduction}

Relying on training from massive datasets to capture extensive common knowledge and having demonstrated certain reasoning capabilities, 
Large Language Models (LLMs) have been widely applied and explored across various domains, rapidly emerging as powerful tools \cite{brown2020language,kojima2022large,tptu,yang2023harnessing}. 
The utilization of prompting techniques, such as chain-of-thought (CoT)~\cite{wei2022chain}, has played a pivotal role in further augmenting the reasoning and planning capabilities of LLMs. This approach eliminates the need for training from scratch by providing an acceptable initial strategy based on common knowledge. 
Examples of such applications include question-answering systems \cite{mallen2023not}, common-sense reasoning \cite{hao2023reasoning}, programming \cite{tian2023chatgpt}, and embodied intelligence \cite{driess2023palm}.

Recently, 
in the fields of natural language processing (NLP) and multi-agent systems (MAS), numerous research endeavors are dedicated to exploring the collaborative task-solving potential facilitated by the cooperation of multiple agents grounded in LLMs.
These efforts leverage the role-playing~\cite{li2023camel} and debate~\cite{chan2023chateval} to facilitate synergy and effective coordination among the agents involved.
However, most existing works focus on coordinating a limited number of agents as shown in Table \ref{tab:comparison_ma}.
The application of LLMs for effective coordination in large-scale agent scenarios has received limited attention.
This is attributed mainly to the significant increase in complexity and difficulty associated with applying LLMs to large-scale multi-agent decision-making tasks.

In this paper, we direct our attention to the following key challenges:
(1) As the number of agents increases, the joint action space grows exponentially, amplifying the difficulty of exploration and exploitation in complex MAS. (2) The limitations of LLMs themselves, as highlighted by the issue of hallucinations \cite{zhang2023siren}, can affect the reliability of decision-making. (3) Effectively managing tokens or communication resources presents a substantial challenge in scenarios involving large-scale LLM-based agents. 
We prioritize these challenges due to their inherent and widespread nature in large-scale settings, and the absence of comprehensive solutions. By focusing on these general challenges, we try to contribute insights and solutions that hold broad relevance, offering a foundational framework for tackling intricacies in diverse real-world scenarios. 

\begin{table*}[t]
\centering
\caption{Comprehensive comparison of LLM-based multi-agent methods. All approaches rely on either multi-agent debate or role-playing to accomplish decision-making tasks and solve NLP problems (task solver), or simulate collective behavior (community simulator).}
\label{tab:comparison_ma}
\resizebox{\textwidth}{!}{%
\begin{tabular}{@{}llccc@{}}
\toprule
\multicolumn{1}{c}{\textbf{Type}} &
  \multicolumn{1}{c}{\textbf{Method}} &
  \textbf{Target} &
  \textbf{Agent Configuration} &
  \textbf{Agents Num.} \\ \midrule
\multirow{3}{*}{\begin{tabular}[c]{@{}l@{}}Muti-Agent \\ Debate\end{tabular}} &
  Debate (\citeauthor{du2023improving}) &
  \multirow{3}{*}{Task Solver} &
  2 debaters &
  2 
   \\
 &
  MAD (\citeauthor{liang2023encouraging}) &
   &
  1 judge + 2 debaters &
  3 
   \\
 &
  ChatEval (\citeauthor{chan2023chateval}) &
   &
  multi debaters &
  5 
   \\ \cmidrule(r){1-5}
\multirow{7}{*}{Role Playing} &
  CAMEL (\citeauthor{li2023camel}) &
  \multirow{4}{*}{Task Solver} &
  1 assistant + 1 user &
  2 
   \\
 &
  AgentVerse (\citeauthor{chen2023agentverse}) &
   &
  1 role assigner + 2-4 experts + 1 evaluater &
  6 
   \\
 &
  Proagent (\citeauthor{zhang2023proagent}) &
   &
  2 cooks &
  2 
   \\
 &
  \textbf{LLaMAC (ours)} &
   &
  \textbf{3 critic + 1-50 actors} &
  \textbf{50} 
   \\ \cmidrule(lr){2-5}
 &
  Generative Agents (\citeauthor{park2023generative}) &
  \multirow{3}{*}{\begin{tabular}[c]{@{}c@{}} Community \\ Simulator\end{tabular}}&
  25 agents &
  25 
   \\
 &
  Werewolf Agents (\citeauthor{xu2023exploring}) &
   &
  7 players &
  7 
  \\
 &
  ReCon (\citeauthor{wang2023avalon}) &
   &
  6 players &
  6 
   \\ \bottomrule
\end{tabular}%
}
\end{table*}

To this end, we present \textbf{L}arge \textbf{La}nguage \textbf{M}odel-based \textbf{A}ctor-\textbf{C}ritic (\textbf{LLaMAC}), a novel framework for achieving a comprehensive decision-making process in collaborative tasks involving large-scale LLM-based agents, drawing inspiration from the classical actor-critic reinforcement learning (RL) approach \cite{konda1999actor}. 
Within LLaMAC, we design a centralized critic which takes on the role of a coordinator, making suggestions to each actor based on their decision memory. 
Subsequently, the actors engage in interactions with the environment, receiving assigned tasks, conducting analyses, and performing corresponding actions.
Specifically, our primary contributions are as follows:
\begin{itemize}
    \item To attain a viable and robust initial strategy and tackle the exploration-exploitation trade-off inherent in the decision-making process, we introduce the TripletCritic structure, which is inspired by the distributional code for value in the brain \cite{dabney2020distributional}. This architecture effectively coordinates multiple critics with shared objectives but varying preferences through internal feedback, thereby providing dependable  action suggestion for each actor involved.
    \item We also establish an external feedback mechanism between the LLM-based actors (i.e., agents) and the TripletCritic. This mechanism serves to not only reduce the access cost of the LLM but also allows each actor to maintain independent exploration and decision-making capabilities.
    \item We propose a modular and token-efficient framework for augmenting the decision-making capabilities of LLM-based agents in large-scale multi-agent environments. This framework enables autonomous iterative cooperation among a large number of agents.
\end{itemize}

We first evaluate the performance of our method on a system resource allocation task to demonstrate its ability to strike a balance between exploration and exploitation, as well as its capability in large-scale multi-agent decision-making tasks. We further deploy our method in a more complex robot grid transportation scenario to validate its planning and decision-making capabilities. Experimental results demonstrate that our method outperforms existing approaches in terms of final performance, token utilization efficiency, and policy stability. To the best of our knowledge, we are the first to apply LLMs to large-scale multi-agent decision-making tasks involving more than 50 agents, as indicated in Table~\ref{tab:comparison_ma}.

\begin{figure*}[t]
    \centering
    \includegraphics[width=6.0 in]{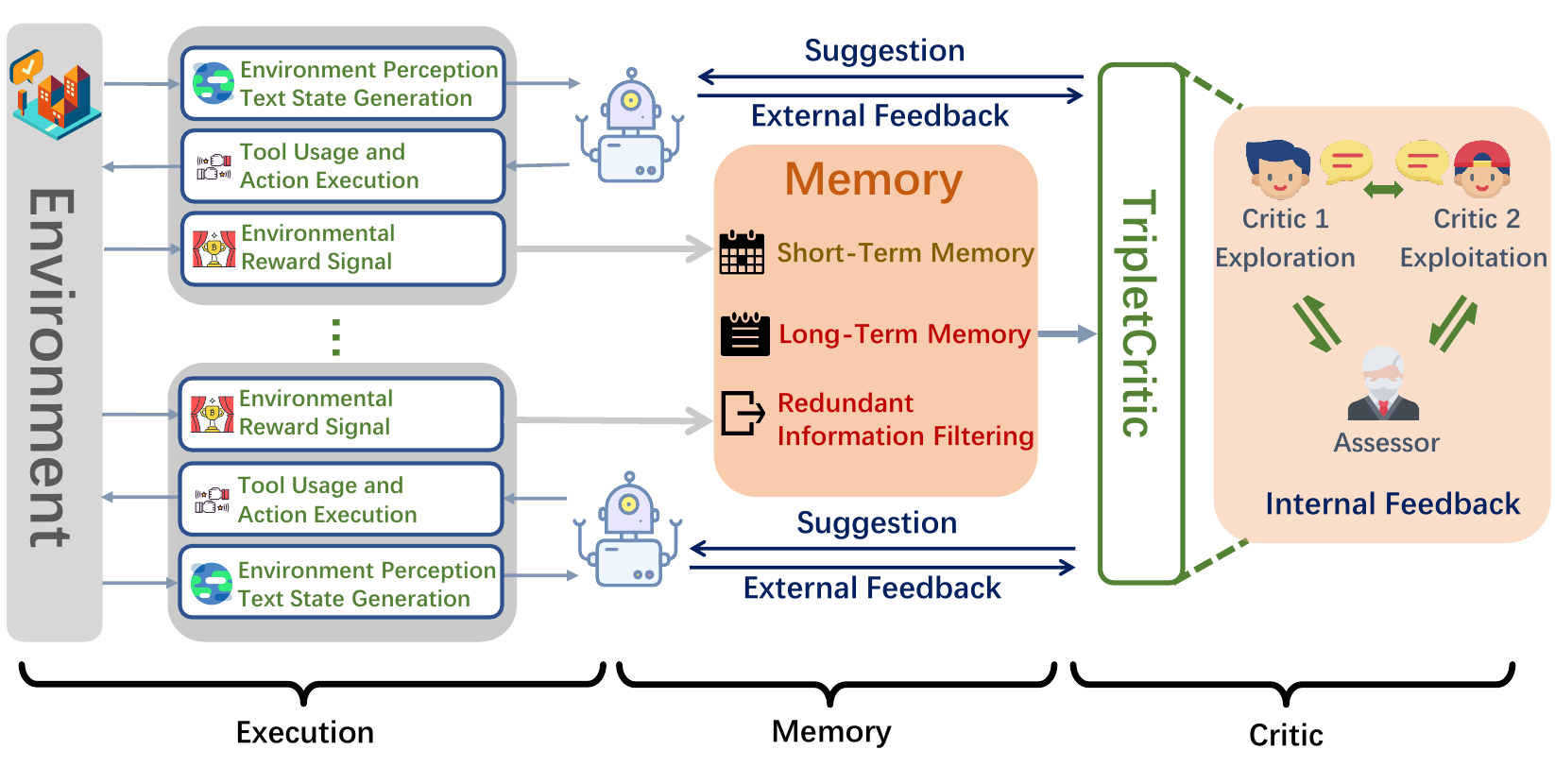}
    \caption{The overall framework of LLaMAC. The LLM-based agents achieve autonomous and continuous decision-making and interaction through the utilization of the execution, memory, and critic modules.}
    \label{fig:Mechanism}
    \vspace{-8pt}
\end{figure*}

\section{Related Work}

\subsection{Multi-Agent Cooperation} 
Extensive research has been conducted to explore collaborative control among agents in MAS, with the objective of acquiring optimal strategies to accomplish ultimate goals. Game theory and RL serve as essential theoretical and practical foundations for this research \cite{yang2020overview,zhang2021multi} \nocite{mao2023pdit,zhang2022efficient}, leading to the development of several novel collaborative training frameworks that effectively address challenges such as equilibrium strategy solving \cite{kuba2021trust,zhang2023inducing,zhang2023stackelberg}, credit assignment \cite{zhou2020learning}, non-stationarity of the environment, and partial observability \cite{rashid2020monotonic}\nocite{xu2023consensus}. Among these approaches, the Actor-Critic method \cite{lowe2017multi}, widely recognized as one of the classical RL techniques, has found extensive application within the context of MAS. Within this framework, a centralized critic estimates the value function to evaluate the quality of policies, while decentralized actors employ gradient ascent based on these assessments to improve their policies, thereby maximizing the expected cumulative return.
However, these methods often suffer from limitations in generalization and require exploration of a large number of irrelevant trajectories, resulting in low training efficiency. Moreover, strategies generated by such black-box optimization methods often lack interpretability. In contrast, our approach enables optimal strategy formulation through a stable and efficient framework based on natural language interaction, providing a transparent and interpretable decision-making process.

\subsection{Planning and Reasoning with LLM}

Learning in massive corpora gives LLMs certain commonsense reasoning capabilities \cite{kojima2022large}. Although there are still challenges in solving complex decision tasks, a large amount of work has proven that their methods can effectively improve the planning ability of LLMs \cite{zelikman2022star,creswell2022selection}. One line of research focuses on decomposing complex queries into sequential intermediate steps, known as Chain-of-Thought (CoT) \cite{wei2022chain}, to achieve accurate solutions. Another direction involves incorporating feedback mechanisms, showcasing their extensive capabilities in tackling complex decision-making challenges \cite{wang2023mint}.
Moreover, recent studies have begun to address this issue employing multiple LLMs.
These approached are enhanced in their planning capabilities through techniques such as debate \cite{chan2023chateval,liang2023encouraging} or role-playing \cite{li2023camel,hong2023metagpt}.
In the domain of decision-making, a subset of research utilizes prompting techniques to construct comprehensive processes covering perception, planning, and action, in  cluding video games \cite{zhang2023proagent}, robot contro l \cite{zhang2023building,mandi2023roco}, and open-world tasks \cite{zhu2023ghost,gong2023mindagent}. There are some studies about task planning and external tool usage \cite{ruan2023tptu-ws,tptu-v2}. 
However, it is worth noting that the existing studies, as outlined in Table~\ref{tab:comparison_ma}, have predominantly concentrated on tasks involving a limited number of agents. The involvement of a larger number of agents has primarily been observed in methods that analyze community simulations, wherein task-solving is not a requirement. In light of this observation, our work uniquely emphasizes the application of language models in the realm of decision-making within large-scale multi-agent systems.

\section{LLaMAC}
In this section, we formally present a systematic and modular framework designed for LLM-based agents, 
namely Large Language Model-based Actor-Critic (LLaMAC), 
with a specific emphasis on their suitability for large-scale decision-making contexts. 

\subsection{Problem Formulation}
This study focuses on the collaborative task solving of MAS, which can be formalized as a Goal-Augmented Decentralized Partially Observable Markov Decision Process (GA-Dec-POMDP) \cite{spaan2012partially}. It is defined by a tuple: $\Gamma\triangleq\langle\mathcal{I}, \mathcal{S},\mathcal{G}, \{\mathcal{O}^i\}_{i\in\mathcal{I}},\{\mathcal{A}^i\}_{i\in\mathcal{I}}, \mathcal{P}, R\rangle$, 
where $\mathcal{I}$, $\mathcal{S}$, $\mathcal{G}$, $\mathcal{O}$, and $\mathcal{A}$ represent the sets of agents, state space, goal space, observation space, and action space, respectively. 
$\mathcal{P}: \mathcal{S} \times \mathcal{A} \times \mathcal{S} \rightarrow [0,1]$ denotes the dynamic transition function, and $\mathcal{R}: \mathcal{S} \times \mathcal{A} \rightarrow \mathbb{R}$ represents the reward function. 
Within a given state $s \in \mathcal{S}$, each agent $i \in \mathcal{I}$ possesses its own local observations $o^i \in \mathcal{O}^i$ within its field of view and performs action $a^i \in \mathcal{A}^i$ accordingly.
Formally, this problem requires each agent $i$ to learn a decision policy $\pi^i:\mathcal{O}^i \rightarrow \mathcal{A}^i$ to solve the task with a goal, which is equivalent to maximizing cumulative rewards.


\subsection{Overall Framework}
As illustrated in Figure~\ref{fig:Mechanism}, LLaMAC introduces the Centralized Critic with Decentralized Actor (CCDA) structure, where actors and critics are LLM-based agents.
The system incorporates three fundamental modules to facilitate a comprehensive decision-making process, enabling iterative reasoning, planning, and continuous interaction between the agents and the environment. 
The functionalities of each module are as follows:

\begin{figure}[t]
    \centering
    \includegraphics[width=3.5in]{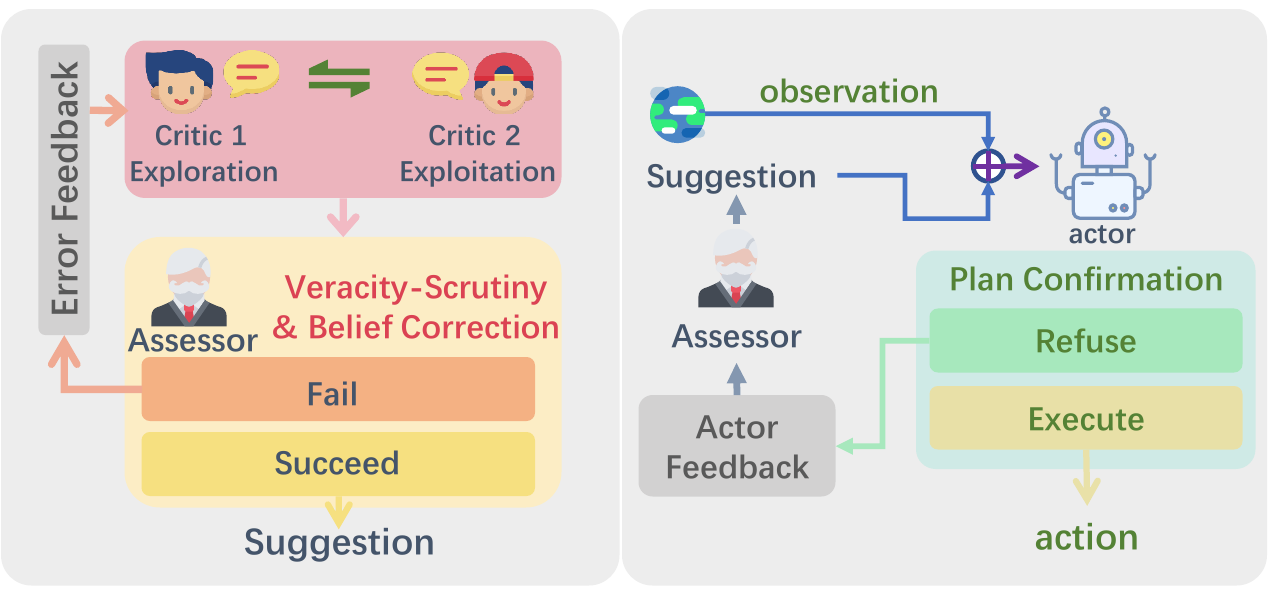}
    \caption{Internal Feedback within the TripletCritic (\emph{Left}) and External Feedback mechanism from actor to critic (\emph{Right}).}
    \label{fig:if}
    \vspace{-12pt}
\end{figure}

\noindent\textbf{Execution Module.}  The execution module fulfills the vital function of converting the original state information obtained from the environment into text-based descriptions that can be comprehended and processed by the language model. The actions performed by each actor encompass a broad spectrum, ranging from intricately detailed actions like adjusting the joint movement angles of a robot to more abstract and higher-level actions such as issuing instructions for the utilization of a specific tool.

\noindent\textbf{Memory Module.} The memory module serves to store crucial information needed during the decision-making process to aid the accumulation of useful knowledge and enhance the agent's decision-making capabilities. 
Specifically, the short-term memory is used to store the most recent state. In contrast, the historical trajectory and experiential information learned from interactions are stored in the long-term memory. The memory module also incorporates a mechanism for filtering redundant information. During long-term planning processes, it retains only the most recent $L$ steps of state transitions $<s_{t-L+1},a_{t-L+1},r_{t-L+1},s_{t-L+2},...,s_{t}>$. This assists the agent in comprehending the relationship between actions and changes in environmental states. 

\noindent\textbf{Critic Module.} 
The critic module assumes a central role within the workflow of LLaMAC.
It receives the present \emph{state} and extracts pertinent details from the memory module, enabling evaluation and learning from the actors' historical trajectories. Functioning as a centralized coordinator, the critic module engages in reasoning and planning activities to formulate potential high-reward and reliable plan suggestions. These suggestions then serve as guides for the interaction between the actor and the environment.

Furthermore, we devise a comprehensive feedback mechanism along with a token-efficient solution to address the challenges posed by the increase in the number of agents,
such as exacerbation of hallucinatory phenomena, escalation of the access cost, and the trade-offs involved in exploration and exploitation.
By coordinating the functionalities of each module and incorporating the feedback mechanism, we have the coherent decision-making workflow:

\begin{itemize}
    \item [(1)] The environment produces a new \emph{state}, denoted as $s$, which is presented in textual format to enable processing by the language model-based agent.
    \item [(2)] The critic receives the state and extracts the relevant information from the memory module. Utilizing these inputs, it facilitates a three-critic dialogue (\emph{Internal Feedback}) and subsequently generates the textual \emph{suggestion} denoted as $su$ for each actor. 
    \item [(3)] Each actor is provided with the \emph{observation} denoted as $o$ from the environment, as well as the \emph{suggestion} $su$ from the TripletCritic. Subsequently, actors engage in a process called \emph{External Feedback}.
    \begin{itemize}
        \item [(3.1)] If all actors reach a consensus that the suggestion is correct, each actor generates an action $a$ based on the information $<o, su>$ and executes the action $a$ in the environment. The environment provides a reward $r$ to the agents, indicating the quality of the action. The entire state transition process is stored in the memory module. Subsequently, a new round of interaction commences, signifying a return to step (1).
         \item [(3.2)] If an actor identifies that the suggestion is incorrect, an external feedback signal is generated. Subsequently, the TripletCritic receives this external feedback information and formulates a new suggestion for the actor based on the three-critic dialogue history and the recently received feedback information. The TripletCritic then transmits the revised suggestion to the respective actor, and the workflow resumes at step (3).
    \end{itemize}
    \item [(4)] The task concludes either when the goal is successfully achieved or when the maximum iteration limit is reached, at which point the final task results are returned.

    
\end{itemize}

\subsection{TripletCritic with Internal Feedback}
\label{sec:internal feedback}



The increasing number of agents presents formidable challenges to the accuracy and efficiency of task evaluation and planning conducted by the critic module. 
The expansion of coordinated action spaces and the growing inter-dependencies in decision-making among agents significantly amplify the complexity of decision-making for language models. 
Moreover, these factors intensify the already challenging issue of hallucinations.

To this end, we develop the TripletCritic, which incorporates an internal feedback mechanism. The design of TripletCritic is inspired by the distributed encoding of reward and value by dopamine neurons in the brain \cite{dabney2020distributional}. Each dopamine neuron contains partial information about the actual reward, and different dopamine neurons utilize different value predictions, enabling the brain to model the value distribution of reward events. 
Similarly, as depicted in Figure~\ref{fig:if}, the TripletCritic framework encompasses a dual-critic structure, each with the same objective but distinct preferences, alongside the third critic, called the assessor, who assumes the responsibility of reconciling these preferences. One critic exhibits a proclivity for exploration, prioritizing long-term gains, while the other gravitates towards exploitation, emphasizing short-term gains. The assessor fulfills two primary roles. Firstly, it makes \emph{Veracity Scrutiny} to check the strategies employed by the dual-critic, offering internal feedback in the event of errors. 
Secondly, it undertakes \emph{Belief Correction} in order to establish a harmonious equilibrium between exploration and exploitation within the planners. 
Additionally, the assessor collaborates with the actors to transmit the final suggestion assignment, informed by these assessments and corrections.

\subsection{External Feedback from Actor to Critic}
\label{sec:external feedback}

\begin{figure*}[t]
    \centering
    \includegraphics[width=6.4 in]{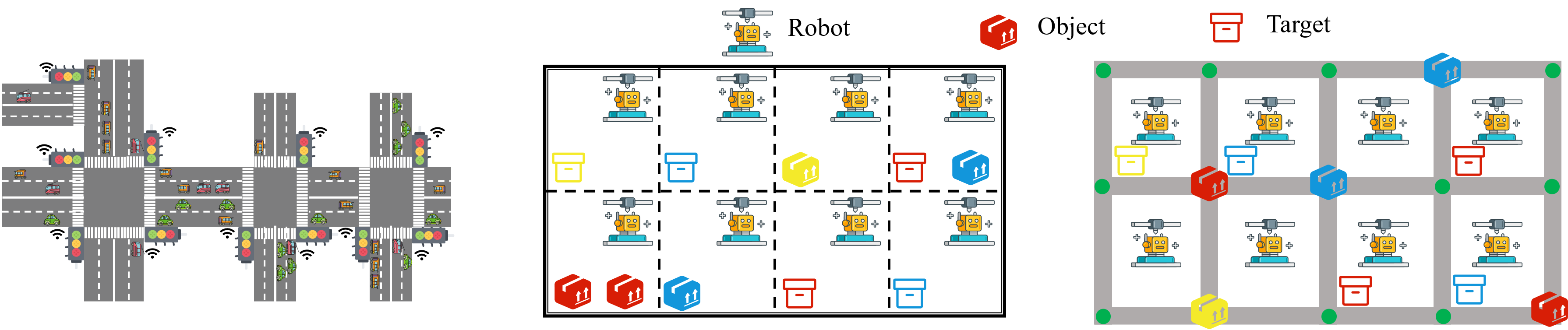}
    \caption{Multi-agent task planning environments. \emph{Left}: System resource allocation, exemplified by addressing traffic congestion. \emph{Middle}: Grid Transportation-Easy. \emph{Right}: Grid Transportation-Hard.}
    \label{fig:env}
    \vspace{-4pt}
\end{figure*}

\begin{figure*}[t]
    \centering
    \includegraphics[width=6.8 in]{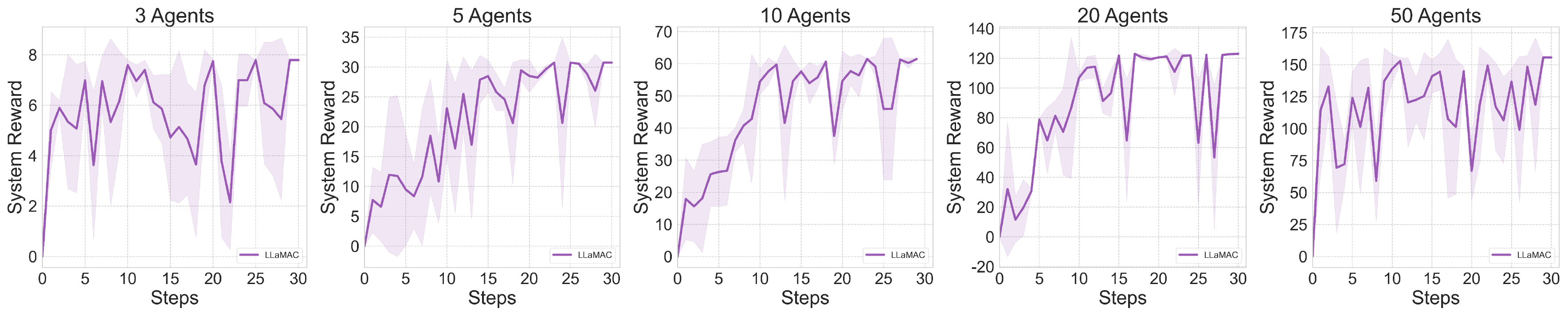}
    \caption{The evaluation performance of LLaMAC in system resource allocation scenarios with different number of agents.}
    \label{fig:GS_LLMAC_results}
\end{figure*}

The TripletCritic provides each actor with a potential initial feasible solution. To facilitate the iterative long-term planning process and achieve the ultimate goal, as well as to reduce the access costs of decision-making for a large number of intelligent agents, we additionally incorporate an external feedback mechanism from actor to critic.

Initially, as depicted in Figure~\ref{fig:if}, the TripletCritic sends \emph{suggestions} $\{su^i\}_{i\in \mathcal{I}}$ to each actor, and all actors pass the proposed plans through an external \emph{Plan Confirmation} to determine their feasibility. If further improvements are deemed necessary, the corresponding LLM is accessed. The LLM takes as input the agent's \emph{observation} $o^i$ and the corresponding \emph{suggestion} $su^i$, providing insights into the underlying issues and potential enhancement strategies. 
Once feedback is received from all actors, the information is aggregated and sent back to the Assessor within the TripletCritic. The Assessor utilizes the internal feedback dialogue information and the actors' external feedback to further update the suggestions for actors with identified issues, returning new \emph{suggestions} to the respective actors. This iterative process continues until all actors determine that no further improvements are necessary, at which point actions are executed directly. 


The coordination among various modules is facilitated by both internal and external feedback, thus forming a comprehensive and automated iterative planning process. TripletCritic enhances the viability and robustness of the initial policy by incorporating an internal feedback mechanism and an evaluation mechanism that balances different preferences. Additionally, it effectively reduces the occurrence of hallucination issues. It is important to highlight that the reliability of TripletCritic reduces the actors' opportunity to provide external feedback, thereby minimizing access costs and promoting the development of token-efficient solutions. The occasional external feedback process further improves the performance of the ultimate strategy.

\section{Evaluation}
In this section, we employ the state-of-the-art large language model, namely GPT-4 \cite{openai2023gpt4}, to conduct a comprehensive evaluation of the effectiveness of our method within two distinct categories of scenarios, as illustrated in Figure~\ref{fig:env}.
Firstly, we examine system resource allocation scenarios to primarily assess the performance of the TripletCritic. Secondly, we explore robot grid transportation scenarios to showcase the performance of LLaMAC in long-term iterative decision-making throughout the entire process.

\begin{figure*}[t]
    \centering
    \includegraphics[width=6.2 in]{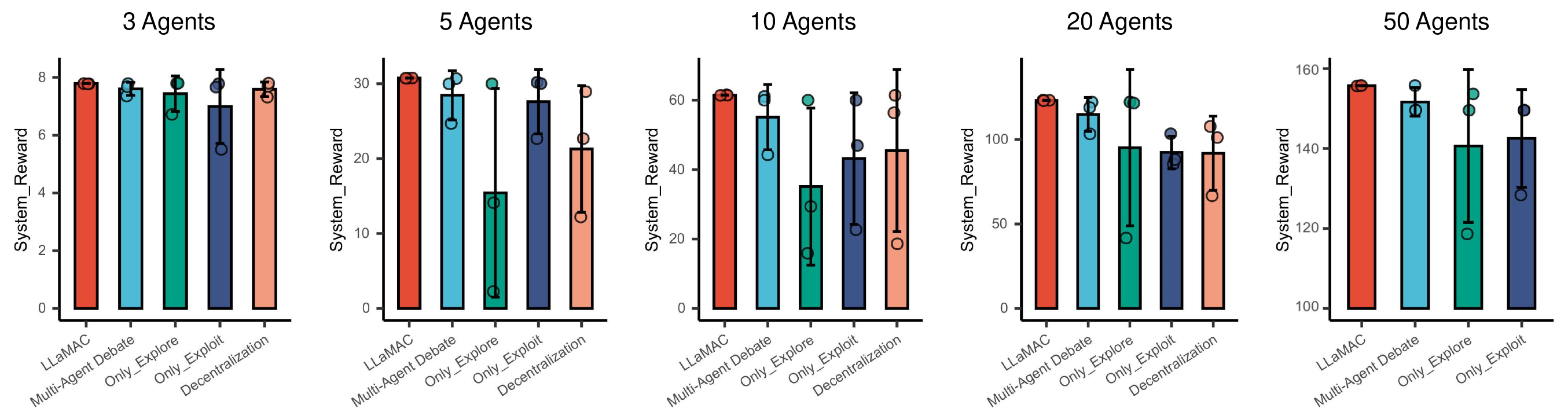}
    \caption{The final performance of different methods in system resource allocation scenarios with different number of agents.}
    \label{fig:GS_final_performance}
    \vspace{-8pt}
\end{figure*}
\begin{figure*}[htbp]
    \centering
    \includegraphics[width=5.7 in]{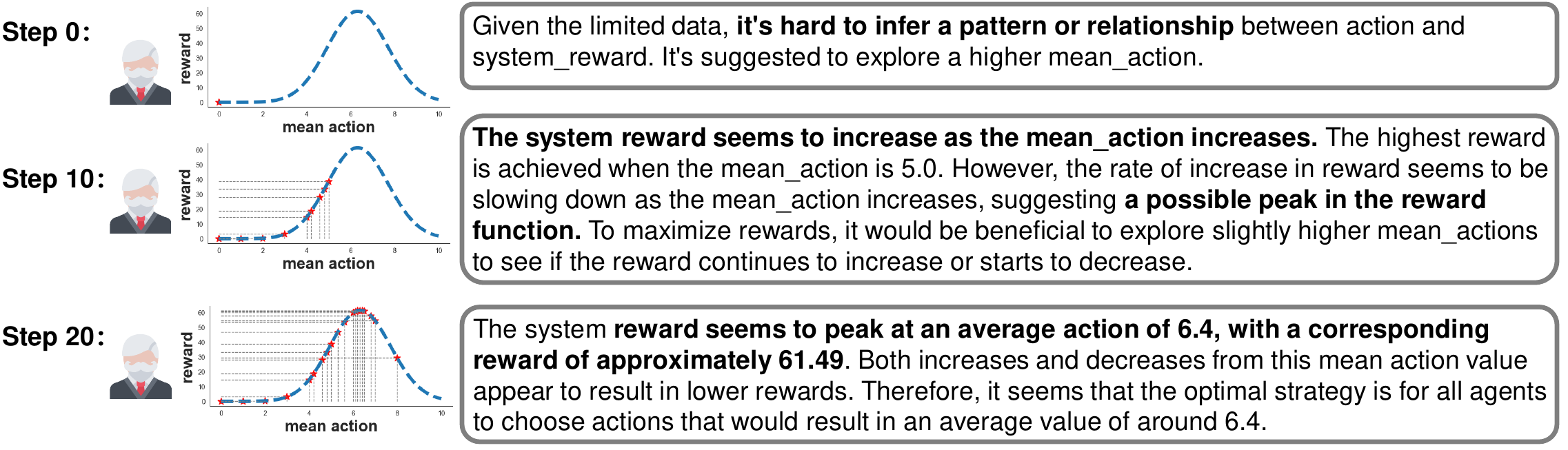}
    \caption{The Assessor in system resource allocation scenario undertakes the crucial tasks of data collection and cognitive analysis. The blue dashed line represents the reward function, while the red dots indicate the explored actions.}
    \label{fig:GS_thought}
    \vspace{-8pt}
\end{figure*}

\subsection{System Resource Allocation}

\subsubsection{Experimental Settings}
System resource allocation \cite{holmesparker2014exploiting} can be viewed as a single-step decision and optimization problem that require mathematical reasoning capabilities of LLMs. It has numerous practical applications, such as addressing traffic congestion. 
In this context, the primary objective is to achieve effective system resource allocation among multiple traffic controllers acting as agents. These agents play a crucial role in directing vehicles onto the main road, optimizing the utilization of the main route while mitigating congestion.

In our experimental setup, the system objective function is defined as the Gaussian squeeze function:
$R(x)=xe^{-\frac{(x-\mu)^2}{\sigma^2}}$,
where $x=\sum_{i\in\mathcal{I}}a^i$ represents the sum of actions chosen by all agents, $\mu$ and $\sigma$ are inherent parameters of the system representing the mean and variance, respectively.

In this scenario, each agent is capable of selecting an integer between 0 and 9 as their action, with no knowledge of the choices made by other agents. The objective for the agents is to synthesize their experiences from multiple decision rounds and infer the allocation scheme that leads to the maximum rewards. Centralized critic possesses the authority to access the actions taken by all agents and the corresponding average values of these actions. 
This particular scenario is highly suited for validating the capabilities of the TripletCritic.

Specifically, we consider scenarios with different numbers of agents, namely 3, 5, 10, 20, and 50. As the number of agents increases, the difficulty of decision-making escalates. We examine several comparative experimental setups, including the \emph{Multi-agent Debate} method \cite{chan2023chateval}, which has recently been utilized in the field of NLP to alleviate hallucinations and enhance mathematical reasoning abilities. Additionally, we explore the \emph{Only\_Explore} approach that solely utilizes a critic biased towards exploration, the \emph{Only\_Exploit} approach that employs a critic biased towards exploitation, and the \emph{Decentralization} method where each agent independently makes decisions based on its own observation history. Due to limitations in terms of access costs, we solely test the \emph{Decentralization} method for scenarios involving fewer than 20 agents.


\subsubsection{Results}

\begin{table*}[t]
\centering
\footnotesize
\setlength{\tabcolsep}{12pt}
\caption{Evaluation results under different grid settings in the Grid Transportation scenarios include metrics such as the success rate (\emph{Success}), time steps (\emph{Steps}) taken to execute tasks, and the count of feedback instances (\emph{Feedback}). The values in parentheses correspond to a single standard deviation over 10 trials.}
\vspace{-8pt}
\label{tab:results_GT}
\begin{tabular}{@{}clcccccc@{}}
\toprule
\multicolumn{1}{l}{} &        & \multicolumn{3}{c|}{Grid Transportation-Easy}                                   & \multicolumn{3}{c}{Grid Transportation-Hard}     \\
                     &        & \textbf{Success}        & \textbf{Steps}      & \multicolumn{1}{c|}{\textbf{Feedback}}             & \textbf{Success}        & \textbf{Steps}      & \textbf{Feedback}    \\ \midrule
\multirow{2}{*}{\textbf{2$\times$2}} & \textbf{HMAS-2} & \textbf{100\%} & 9.9(2.74)  & \multicolumn{1}{c|}{3.3(2.05)}   & 80\%    & 7.0(5.0)  & 6.0(9.74)   \\
 &
  \textbf{LLaMAC} &
  \textbf{100\%} &
  \textbf{7.0(1.79)} &
  \multicolumn{1}{c|}{\textbf{2.0(1.26)}} &
  \textbf{100\%} &
  \textbf{4.7(1.35)} &
  \textbf{3.6(2.80)} \\ \midrule
\multirow{2}{*}{\textbf{2$\times$4}} & \textbf{HMAS-2} & 80\%           & 15.5(6.09)   & \multicolumn{1}{c|}{12.3(5.83)} & 20\%    & 17.0(9.0) & 24.0(20.0) \\
 &
  \textbf{LLaMAC} &
  \textbf{100\%} &
  \textbf{7.6(1.36)} &
  \multicolumn{1}{c|}{\textbf{4.3(1.42)}} &
  \textbf{90\%} &
  \textbf{7.44(2.95)} &
  \textbf{10.56(7.54)}  \\ \midrule
\multirow{2}{*}{\textbf{4$\times$8}} & \textbf{HMAS-2} & 60\%           & 30.6(9.70)  & \multicolumn{1}{c|}{26.1(13.59)} & 0\%     & -         & -              \\
 &
  \textbf{LLaMAC} &
  \textbf{100\%} &
  \textbf{12.9(2.70)} &
  \multicolumn{1}{c|}{\textbf{10.7(3.35)}} &
  \textbf{90\%} &
  \textbf{8.44(1.57)} &
  \textbf{12.11(2.51)} \\ \bottomrule
\end{tabular}%
\vspace{-8pt}
\end{table*}

As shown in Figure~\ref{fig:GS_LLMAC_results}, it is evident that within a limited number of steps, LLaMAC demonstrates the ability to explore and learn through continuous interaction with the environment. The final performance of all methods is presented in Figure~\ref{fig:GS_final_performance}. The TripletCritic approach within LLaMAC exhibits a similar structure to the Multi-agent Debate method, and compared to other approaches, these two methods display relatively stable performance. However, debate-based methods often suffer from excessive or insufficient exploration, resulting in a tendency to converge to local optima. On the other hand, approaches that emphasize exploration and exploitation struggle to maintain stable performance. The former exhibits significant oscillations due to excessive exploration, while the latter prematurely converges to local optima after only a few simple exploratory steps, aligning with the expected characteristics of these methods. 
Distributed approach incurs the highest access cost , as each agent is required to independently access the LLM. Nevertheless, the lack of collaboration among the agents still hinders the capture of true relationships.


\subsubsection{Case Study}
We explicitly depict the cognitive process of the assessor after continuous data collection, as illustrated in  Figure~\ref{fig:GS_thought}. It can be observed that LLaMAC is capable of providing insightful recommendations based on the current state of data collection, aiding in further inference of the relationship between actions and rewards. At step 10, the collected data only reveals a positive correlation between actions and rewards. However, remarkably, the Assessor accurately identifies the non-linear growth pattern of rewards and infers the existence of a potential peak in the objective function. After 20 decision rounds, the Assessor successfully identifies the optimal value and conducts thorough exploration near the peak to avoid getting trapped in local optima.

\subsection{Grid Transportation}

\subsubsection{Experimental Settings}
The robot grid transportation task is relatively more complex as it simulates the automatic control system of robots in factory assembly line operations. It can be considered as a multi-step decision problem that requires the spatial reasoning and logical reasoning capabilities of LLMs. Additionally, it puts the long-term planning ability to the test. We consider two environmental configurations:

\noindent\textbf{Grid Transportation-Easy.} The environment consists of a grid of size $N\times M$, with one intelligent agent assigned to each grid cell. Different types of objects and targets are unevenly distributed across the grid. The objective of the intelligent agents is to transport all objects to their respective targets. The available actions for each agent include moving an object to a horizontally or vertically adjacent grid cell, or placing an object into the target location if both the object and target are in the same grid cell.

\noindent\textbf{Grid Transportation-Hard.} The task goals are the same as in the easy scenario, with the key difference being that objects can only move along the grid boundaries. Each robot's available actions include moving an object located at one of the four corners of its grid cell to one of the other three corners, or to the target location if the object's target position is within the grid.
In this scenario, the interdependent coordination among agents becomes more complex. Objects located at a particular corner may be moved simultaneously by multiple agents, leading to conflicts. Additionally, adjacent agents may attempt to move different objects to the same corner, resulting in collisions.

Our objective is to ensure the smooth execution of tasks and the successful accomplishment of goals by LLM-based agents. When an agent experiences hallucinations that persist beyond the specified iteration limit, the task is deemed unsuccessful. This includes instances where the output grammar format fails to meet the requirements even after reaching the maximum number of iterations, when the dialogue context exceeds the token length limit, and when the decision time steps surpass the designated limit.

\subsubsection{Results}

\begin{figure}[t]
    \centering
    \includegraphics[width=3.2 in]{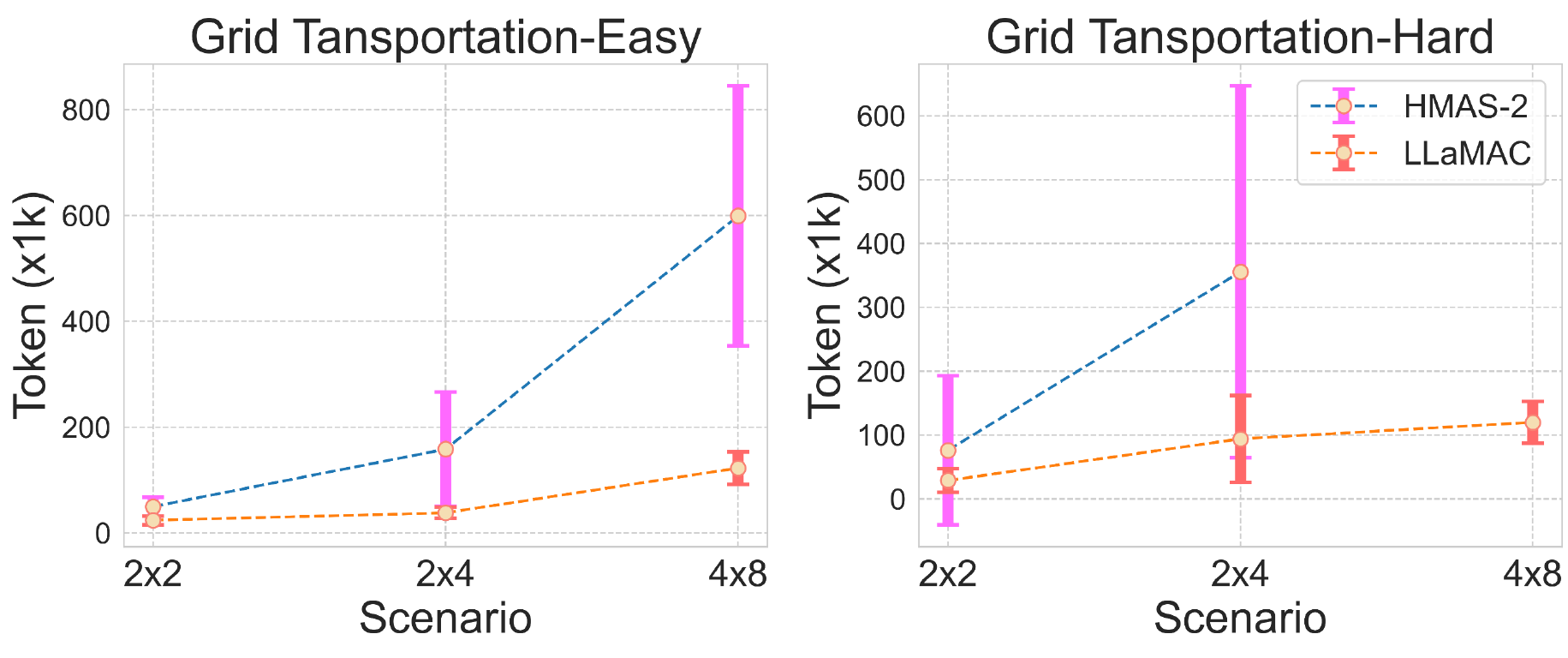}
    \caption{Token usage of LLaMAC and HMAS-2 in the Grid Transportation scenarios.}
    \label{fig:token_effi}
    \vspace{-12pt}
\end{figure}

We conduct a comparative analysis between our method and the state-of-the-art solution, HMAS-2 \cite{chen2023scalable}. For each scenario, we conduct tests on grid configurations of $2\times2$, $2\times4$, and $4\times8$, respectively.
Table~\ref{tab:results_GT} presents a comprehensive performance comparison between the two methods, clearly demonstrating the overall superiority of our approach. In complex scenarios involving long-term iterative decision-making, LLaMAC exhibits a significantly higher success rate compared to HMAS-2. Furthermore, LLaMAC consistently achieves task completion in fewer interaction steps, highlighting the performance advantages of its employed strategies. Additionally, as shown in Figure~\ref{fig:token_effi} the TripletCritic facilitates the generation of superior initial suggestions, thereby reducing the need for feedback iterations and greatly enhancing token utilization efficiency.

\subsubsection{Case Study}
\begin{figure}[t]
    \centering
    \includegraphics[width=3.3 in]{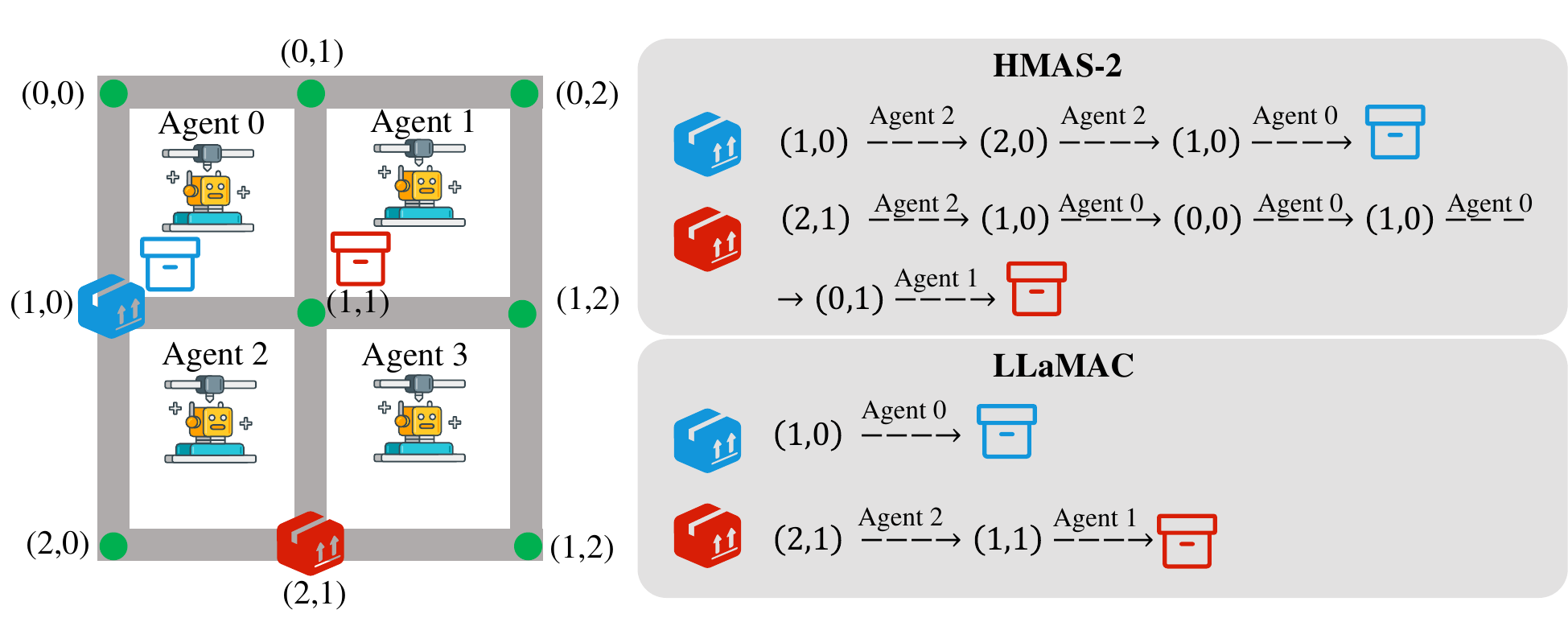}
    \caption{The performance of LLaMAC and HMAS-2 in the 2x2 robotic grid transportation scenario. To enhance visualization, non-essential objects and targets within the scene are concealed.}
    \label{fig:GT_case_study}
    \vspace{-10pt}
\end{figure}

During the experimental process, we observe that LLaMAC effectively enhances the capabilities of LLM in long-term planning and execution, spatial reasoning, and learning from interactions or errors. For example, spatial reasoning poses a significant challenge for LLMs, as they are more prone to hallucinations when determining whether an object is closer to the target. This issue becomes more pronounced in the Hard scenario. As shown in Figure~\ref{fig:GT_case_study}, in the HMAS-2 method, agents often move objects to positions far from the target and may repeatedly move them between two particular locations. In contrast, in LLaMAC, such occurrences are often corrected during the external feedback phase. The actor only needs to focus on its own task, and when it receives suggestions from the critic, the difficulty of determining the effectiveness of individual agent tasks is significantly reduced compared to joint policies. This makes spatial reasoning errors more easily detected, reflected and corrected.

\section{Conclusion}
In this study, we present a novel framework called LLaMAC to enhance the collaborative performance of large-scale multi-agent systems based on Large Language Models. Building upon the commonsense reasoning capabilities exhibited by LLMs, we effectively augment the planning and coordination abilities among agents through stable reasoning mechanisms and comprehensive feedback mechanisms, facilitating continuous interaction between agents and the environment. LLaMAC demonstrates remarkable performance in coordinated scenarios involving a large number of agents. Notably, it exhibits exceptional capabilities in long-term planning, mathematical reasoning and optimization problems, spatial reasoning, and learning from mistakes. Additionally, LLaMAC reduces the access costs associated with large-scale multi-agent collaboration.
We believe that with further enhancements in LLMs and the emergence of more collaboration frameworks, the field of multi-agent collaboration will experience new opportunities for advancement.

\bibliographystyle{named}
\bibliography{ijcai24}

\begin{thebibliography}{}

\bibitem[\protect\citeauthoryear{Brown \bgroup \em et al.\egroup }{2020}]{brown2020language}
Tom Brown, Benjamin Mann, Nick Ryder, Melanie Subbiah, Jared~D Kaplan, Prafulla Dhariwal, Arvind Neelakantan, Pranav Shyam, Girish Sastry, Amanda Askell, et~al.
\newblock Language models are few-shot learners.
\newblock {\em Advances in neural information processing systems}, 33:1877--1901, 2020.

\bibitem[\protect\citeauthoryear{Chan \bgroup \em et al.\egroup }{2023}]{chan2023chateval}
Chi-Min Chan, Weize Chen, Yusheng Su, Jianxuan Yu, Wei Xue, Shanghang Zhang, Jie Fu, and Zhiyuan Liu.
\newblock Chateval: Towards better llm-based evaluators through multi-agent debate.
\newblock {\em arXiv preprint arXiv:2308.07201}, 2023.

\bibitem[\protect\citeauthoryear{Chen \bgroup \em et al.\egroup }{2023a}]{chen2023agentverse}
Weize Chen, Yusheng Su, Jingwei Zuo, Cheng Yang, Chenfei Yuan, Chen Qian, Chi-Min Chan, Yujia Qin, Yaxi Lu, Ruobing Xie, et~al.
\newblock Agentverse: Facilitating multi-agent collaboration and exploring emergent behaviors in agents.
\newblock {\em arXiv preprint arXiv:2308.10848}, 2023.

\bibitem[\protect\citeauthoryear{Chen \bgroup \em et al.\egroup }{2023b}]{chen2023scalable}
Yongchao Chen, Jacob Arkin, Yang Zhang, Nicholas Roy, and Chuchu Fan.
\newblock Scalable multi-robot collaboration with large language models: Centralized or decentralized systems?
\newblock {\em arXiv preprint arXiv:2309.15943}, 2023.

\bibitem[\protect\citeauthoryear{Creswell \bgroup \em et al.\egroup }{2022}]{creswell2022selection}
Antonia Creswell, Murray Shanahan, and Irina Higgins.
\newblock Selection-inference: Exploiting large language models for interpretable logical reasoning.
\newblock {\em arXiv preprint arXiv:2205.09712}, 2022.

\bibitem[\protect\citeauthoryear{Dabney \bgroup \em et al.\egroup }{2020}]{dabney2020distributional}
Will Dabney, Zeb Kurth-Nelson, Naoshige Uchida, Clara~Kwon Starkweather, Demis Hassabis, R{\'e}mi Munos, and Matthew Botvinick.
\newblock A distributional code for value in dopamine-based reinforcement learning.
\newblock {\em Nature}, 577(7792):671--675, 2020.

\bibitem[\protect\citeauthoryear{Driess \bgroup \em et al.\egroup }{2023}]{driess2023palm}
Danny Driess, Fei Xia, Mehdi~SM Sajjadi, Corey Lynch, Aakanksha Chowdhery, Brian Ichter, Ayzaan Wahid, Jonathan Tompson, Quan Vuong, Tianhe Yu, et~al.
\newblock Palm-e: An embodied multimodal language model.
\newblock {\em arXiv preprint arXiv:2303.03378}, 2023.

\bibitem[\protect\citeauthoryear{Du \bgroup \em et al.\egroup }{2023}]{du2023improving}
Yilun Du, Shuang Li, Antonio Torralba, Joshua~B Tenenbaum, and Igor Mordatch.
\newblock Improving factuality and reasoning in language models through multiagent debate.
\newblock {\em arXiv preprint arXiv:2305.14325}, 2023.

\bibitem[\protect\citeauthoryear{Gong \bgroup \em et al.\egroup }{2023}]{gong2023mindagent}
Ran Gong, Qiuyuan Huang, Xiaojian Ma, Hoi Vo, Zane Durante, Yusuke Noda, Zilong Zheng, Song-Chun Zhu, Demetri Terzopoulos, Li~Fei-Fei, et~al.
\newblock Mindagent: Emergent gaming interaction.
\newblock {\em arXiv preprint arXiv:2309.09971}, 2023.

\bibitem[\protect\citeauthoryear{Hao \bgroup \em et al.\egroup }{2023}]{hao2023reasoning}
Shibo Hao, Yi~Gu, Haodi Ma, Joshua~Jiahua Hong, Zhen Wang, Daisy~Zhe Wang, and Zhiting Hu.
\newblock Reasoning with language model is planning with world model.
\newblock {\em arXiv preprint arXiv:2305.14992}, 2023.

\bibitem[\protect\citeauthoryear{HolmesParker \bgroup \em et al.\egroup }{2014}]{holmesparker2014exploiting}
Chris HolmesParker, M~Taylor, Yusen Zhan, and Kagan Tumer.
\newblock Exploiting structure and agent-centric rewards to promote coordination in large multiagent systems.
\newblock In {\em Adaptive and learning agents workshop}, 2014.

\bibitem[\protect\citeauthoryear{Hong \bgroup \em et al.\egroup }{2023}]{hong2023metagpt}
Sirui Hong, Xiawu Zheng, Jonathan Chen, Yuheng Cheng, Ceyao Zhang, Zili Wang, Steven Ka~Shing Yau, Zijuan Lin, Liyang Zhou, Chenyu Ran, et~al.
\newblock Metagpt: Meta programming for multi-agent collaborative framework.
\newblock {\em arXiv preprint arXiv:2308.00352}, 2023.

\bibitem[\protect\citeauthoryear{Kojima \bgroup \em et al.\egroup }{2022}]{kojima2022large}
Takeshi Kojima, Shixiang~Shane Gu, Machel Reid, Yutaka Matsuo, and Yusuke Iwasawa.
\newblock Large language models are zero-shot reasoners.
\newblock {\em Advances in neural information processing systems}, 35:22199--22213, 2022.

\bibitem[\protect\citeauthoryear{Konda and Tsitsiklis}{1999}]{konda1999actor}
Vijay Konda and John Tsitsiklis.
\newblock Actor-critic algorithms.
\newblock {\em Advances in neural information processing systems}, 12, 1999.

\bibitem[\protect\citeauthoryear{Kong \bgroup \em et al.\egroup }{2023}]{tptu-v2}
Yilun Kong, Jingqing Ruan, Yihong Chen, Bin Zhang, Tianpeng Bao, Shiwei Shi, Guoqing Du, Xiaoru Hu, Hangyu Mao, Ziyue Li, Xingyu Zeng, and Rui Zhao.
\newblock Tptu-v2: Boosting task planning and tool usage of large language model-based agents in real-world systems.
\newblock {\em arXiv preprint arXiv:2311.11315}, 2023.

\bibitem[\protect\citeauthoryear{Kuba \bgroup \em et al.\egroup }{2021}]{kuba2021trust}
Jakub~Grudzien Kuba, Ruiqing Chen, Muning Wen, Ying Wen, Fanglei Sun, Jun Wang, and Yaodong Yang.
\newblock Trust region policy optimisation in multi-agent reinforcement learning.
\newblock {\em arXiv preprint arXiv:2109.11251}, 2021.

\bibitem[\protect\citeauthoryear{Li \bgroup \em et al.\egroup }{2023}]{li2023camel}
Guohao Li, Hasan Abed Al~Kader Hammoud, Hani Itani, Dmitrii Khizbullin, and Bernard Ghanem.
\newblock Camel: Communicative agents for "mind" exploration of large language model society.
\newblock In {\em Thirty-seventh Conference on Neural Information Processing Systems}, 2023.

\bibitem[\protect\citeauthoryear{Liang \bgroup \em et al.\egroup }{2023}]{liang2023encouraging}
Tian Liang, Zhiwei He, Wenxiang Jiao, Xing Wang, Yan Wang, Rui Wang, Yujiu Yang, Zhaopeng Tu, and Shuming Shi.
\newblock Encouraging divergent thinking in large language models through multi-agent debate.
\newblock {\em arXiv preprint arXiv:2305.19118}, 2023.

\bibitem[\protect\citeauthoryear{Lowe \bgroup \em et al.\egroup }{2017}]{lowe2017multi}
Ryan Lowe, Yi~I Wu, Aviv Tamar, Jean Harb, OpenAI Pieter~Abbeel, and Igor Mordatch.
\newblock Multi-agent actor-critic for mixed cooperative-competitive environments.
\newblock {\em Advances in neural information processing systems}, 30, 2017.

\bibitem[\protect\citeauthoryear{Mallen \bgroup \em et al.\egroup }{2023}]{mallen2023not}
Alex Mallen, Akari Asai, Victor Zhong, Rajarshi Das, Daniel Khashabi, and Hannaneh Hajishirzi.
\newblock When not to trust language models: Investigating effectiveness of parametric and non-parametric memories.
\newblock In {\em Proceedings of the 61st Annual Meeting of the Association for Computational Linguistics (Volume 1: Long Papers)}, pages 9802--9822, 2023.

\bibitem[\protect\citeauthoryear{Mandi \bgroup \em et al.\egroup }{2023}]{mandi2023roco}
Zhao Mandi, Shreeya Jain, and Shuran Song.
\newblock Roco: Dialectic multi-robot collaboration with large language models.
\newblock {\em arXiv preprint arXiv:2307.04738}, 2023.

\bibitem[\protect\citeauthoryear{Mao \bgroup \em et al.\egroup }{2023}]{mao2023pdit}
Hangyu Mao, Rui Zhao, Ziyue Li, Zhiwei Xu, Hao Chen, Yiqun Chen, Bin Zhang, Zhen Xiao, Junge Zhang, and Jiangjin Yin.
\newblock Pdit: Interleaving perception and decision-making transformers for deep reinforcement learning.
\newblock {\em arXiv preprint arXiv:2312.15863}, 2023.

\bibitem[\protect\citeauthoryear{OpenAI}{2023}]{openai2023gpt4}
OpenAI.
\newblock Gpt-4 technical report, 2023.

\bibitem[\protect\citeauthoryear{Park \bgroup \em et al.\egroup }{2023}]{park2023generative}
Joon~Sung Park, Joseph O'Brien, Carrie~Jun Cai, Meredith~Ringel Morris, Percy Liang, and Michael~S Bernstein.
\newblock Generative agents: Interactive simulacra of human behavior.
\newblock In {\em Proceedings of the 36th Annual ACM Symposium on User Interface Software and Technology}, pages 1--22, 2023.

\bibitem[\protect\citeauthoryear{Rashid \bgroup \em et al.\egroup }{2020}]{rashid2020monotonic}
Tabish Rashid, Mikayel Samvelyan, Christian~Schroeder De~Witt, Gregory Farquhar, Jakob Foerster, and Shimon Whiteson.
\newblock Monotonic value function factorisation for deep multi-agent reinforcement learning.
\newblock {\em The Journal of Machine Learning Research}, 21(1):7234--7284, 2020.

\bibitem[\protect\citeauthoryear{Ruan \bgroup \em et al.\egroup }{2023a}]{tptu}
Jingqing Ruan, Yihong Chen, Bin Zhang, Zhiwei Xu, Tianpeng Bao, Guoqing Du, Shiwei Shi, Hangyu Mao, Xingyu Zeng, and Rui Zhao.
\newblock Tptu: Task planning and tool usage of large language model-based ai agents.
\newblock {\em arXiv preprint arXiv:2308.03427}, 2023.

\bibitem[\protect\citeauthoryear{Ruan \bgroup \em et al.\egroup }{2023b}]{ruan2023tptu-ws}
Jingqing Ruan, YiHong Chen, Bin Zhang, Zhiwei Xu, Tianpeng Bao, Hangyu Mao, Xingyu Zeng, Rui Zhao, et~al.
\newblock Tptu: Task planning and tool usage of large language model-based ai agents.
\newblock In {\em NeurIPS 2023 Foundation Models for Decision Making Workshop}, 2023.

\bibitem[\protect\citeauthoryear{Spaan}{2012}]{spaan2012partially}
Matthijs~TJ Spaan.
\newblock Partially observable markov decision processes.
\newblock In {\em Reinforcement learning: State-of-the-art}, pages 387--414. Springer, 2012.

\bibitem[\protect\citeauthoryear{Tian \bgroup \em et al.\egroup }{2023}]{tian2023chatgpt}
Haoye Tian, Weiqi Lu, Tsz~On Li, Xunzhu Tang, Shing-Chi Cheung, Jacques Klein, and Tegawend{\'e}~F Bissyand{\'e}.
\newblock Is chatgpt the ultimate programming assistant--how far is it?
\newblock {\em arXiv preprint arXiv:2304.11938}, 2023.

\bibitem[\protect\citeauthoryear{Wang \bgroup \em et al.\egroup }{2023a}]{wang2023avalon}
Shenzhi Wang, Chang Liu, Zilong Zheng, Siyuan Qi, Shuo Chen, Qisen Yang, Andrew Zhao, Chaofei Wang, Shiji Song, and Gao Huang.
\newblock Avalon's game of thoughts: Battle against deception through recursive contemplation.
\newblock {\em arXiv preprint arXiv:2310.01320}, 2023.

\bibitem[\protect\citeauthoryear{Wang \bgroup \em et al.\egroup }{2023b}]{wang2023mint}
Xingyao Wang, Zihan Wang, Jiateng Liu, Yangyi Chen, Lifan Yuan, Hao Peng, and Heng Ji.
\newblock Mint: Evaluating llms in multi-turn interaction with tools and language feedback.
\newblock {\em arXiv preprint arXiv:2309.10691}, 2023.

\bibitem[\protect\citeauthoryear{Wei \bgroup \em et al.\egroup }{2022}]{wei2022chain}
Jason Wei, Xuezhi Wang, Dale Schuurmans, Maarten Bosma, Fei Xia, Ed~Chi, Quoc~V Le, Denny Zhou, et~al.
\newblock Chain-of-thought prompting elicits reasoning in large language models.
\newblock {\em Advances in Neural Information Processing Systems}, 35:24824--24837, 2022.

\bibitem[\protect\citeauthoryear{Xu \bgroup \em et al.\egroup }{2023a}]{xu2023exploring}
Yuzhuang Xu, Shuo Wang, Peng Li, Fuwen Luo, Xiaolong Wang, Weidong Liu, and Yang Liu.
\newblock Exploring large language models for communication games: An empirical study on werewolf.
\newblock {\em arXiv preprint arXiv:2309.04658}, 2023.

\bibitem[\protect\citeauthoryear{Xu \bgroup \em et al.\egroup }{2023b}]{xu2023consensus}
Zhiwei Xu, Bin Zhang, Dapeng Li, Zeren Zhang, Guangchong Zhou, Hao Chen, and Guoliang Fan.
\newblock Consensus learning for cooperative multi-agent reinforcement learning.
\newblock In {\em Proceedings of the AAAI Conference on Artificial Intelligence}, volume~37, pages 11726--11734, 2023.

\bibitem[\protect\citeauthoryear{Yang and Wang}{2020}]{yang2020overview}
Yaodong Yang and Jun Wang.
\newblock An overview of multi-agent reinforcement learning from game theoretical perspective.
\newblock {\em arXiv preprint arXiv:2011.00583}, 2020.

\bibitem[\protect\citeauthoryear{Yang \bgroup \em et al.\egroup }{2023}]{yang2023harnessing}
Jingfeng Yang, Hongye Jin, Ruixiang Tang, Xiaotian Han, Qizhang Feng, Haoming Jiang, Bing Yin, and Xia Hu.
\newblock Harnessing the power of llms in practice: A survey on chatgpt and beyond.
\newblock {\em arXiv preprint arXiv:2304.13712}, 2023.

\bibitem[\protect\citeauthoryear{Zelikman \bgroup \em et al.\egroup }{2022}]{zelikman2022star}
Eric Zelikman, Yuhuai Wu, Jesse Mu, and Noah Goodman.
\newblock Star: Bootstrapping reasoning with reasoning.
\newblock {\em Advances in Neural Information Processing Systems}, 35:15476--15488, 2022.

\bibitem[\protect\citeauthoryear{Zhang \bgroup \em et al.\egroup }{2021}]{zhang2021multi}
Kaiqing Zhang, Zhuoran Yang, and Tamer Ba{\c{s}}ar.
\newblock Multi-agent reinforcement learning: A selective overview of theories and algorithms.
\newblock {\em Handbook of reinforcement learning and control}, pages 321--384, 2021.

\bibitem[\protect\citeauthoryear{Zhang \bgroup \em et al.\egroup }{2022}]{zhang2022efficient}
Bin Zhang, Yunpeng Bai, Zhiwei Xu, Dapeng Li, and Guoliang Fan.
\newblock Efficient policy generation in multi-agent systems via hypergraph neural network.
\newblock In {\em International Conference on Neural Information Processing}, pages 219--230. Springer, 2022.

\bibitem[\protect\citeauthoryear{Zhang \bgroup \em et al.\egroup }{2023a}]{zhang2023inducing}
Bin Zhang, Lijuan Li, Zhiwei Xu, Dapeng Li, and Guoliang Fan.
\newblock Inducing stackelberg equilibrium through spatio-temporal sequential decision-making in multi-agent reinforcement learning.
\newblock {\em arXiv preprint arXiv:2304.10351}, 2023.

\bibitem[\protect\citeauthoryear{Zhang \bgroup \em et al.\egroup }{2023b}]{zhang2023stackelberg}
Bin Zhang, Hangyu Mao, Lijuan Li, Zhiwei Xu, Dapeng Li, Rui Zhao, and Guoliang Fan.
\newblock Stackelberg decision transformer for asynchronous action coordination in multi-agent systems.
\newblock {\em arXiv preprint arXiv:2305.07856}, 2023.

\bibitem[\protect\citeauthoryear{Zhang \bgroup \em et al.\egroup }{2023c}]{zhang2023proagent}
Ceyao Zhang, Kaijie Yang, Siyi Hu, Zihao Wang, Guanghe Li, Yihang Sun, Cheng Zhang, Zhaowei Zhang, Anji Liu, Song-Chun Zhu, et~al.
\newblock Proagent: Building proactive cooperative ai with large language models.
\newblock {\em arXiv preprint arXiv:2308.11339}, 2023.

\bibitem[\protect\citeauthoryear{Zhang \bgroup \em et al.\egroup }{2023d}]{zhang2023building}
Hongxin Zhang, Weihua Du, Jiaming Shan, Qinhong Zhou, Yilun Du, Joshua~B Tenenbaum, Tianmin Shu, and Chuang Gan.
\newblock Building cooperative embodied agents modularly with large language models.
\newblock {\em arXiv preprint arXiv:2307.02485}, 2023.

\bibitem[\protect\citeauthoryear{Zhang \bgroup \em et al.\egroup }{2023e}]{zhang2023siren}
Yue Zhang, Yafu Li, Leyang Cui, Deng Cai, Lemao Liu, Tingchen Fu, Xinting Huang, Enbo Zhao, Yu~Zhang, Yulong Chen, et~al.
\newblock Siren's song in the ai ocean: A survey on hallucination in large language models.
\newblock {\em arXiv preprint arXiv:2309.01219}, 2023.

\bibitem[\protect\citeauthoryear{Zhou \bgroup \em et al.\egroup }{2020}]{zhou2020learning}
Meng Zhou, Ziyu Liu, Pengwei Sui, Yixuan Li, and Yuk~Ying Chung.
\newblock Learning implicit credit assignment for cooperative multi-agent reinforcement learning.
\newblock {\em Advances in neural information processing systems}, 33:11853--11864, 2020.

\bibitem[\protect\citeauthoryear{Zhu \bgroup \em et al.\egroup }{2023}]{zhu2023ghost}
Xizhou Zhu, Yuntao Chen, Hao Tian, Chenxin Tao, Weijie Su, Chenyu Yang, Gao Huang, Bin Li, Lewei Lu, Xiaogang Wang, et~al.
\newblock Ghost in the minecraft: Generally capable agents for open-world enviroments via large language models with text-based knowledge and memory.
\newblock {\em arXiv preprint arXiv:2305.17144}, 2023.

\end{thebibliography}

\newpage
\onecolumn
\appendix
\numberwithin{equation}{section}
\numberwithin{figure}{section}
\numberwithin{table}{section}
\renewcommand{\thesection}{{\Alph{section}}}
\renewcommand{\thesubsection}{\Alph{section}.\arabic{subsection}}
\renewcommand{\thesubsubsection}{\Roman{section}.\arabic{subsection}.\arabic{subsubsection}}
\setcounter{secnumdepth}{-1}
\setcounter{secnumdepth}{3}

\section{Implementation Details}

\subsection{Pseudo-Code}
\label{sec:llamac_pc}
\begin{algorithm}[htbp]
\caption{Execution Procedure for LLaMAC}
\textbf{Hyperparameters}: Length of episode $T$, number of agents $N$, trajectory length of Memory $L$, maximum number of internal and external feedback iterations $IF, EF$\\
\textbf{Initialize}: Memory $\mathcal{M}$, Environmental initial state $s_0$ and observation $\{o_0^i\}_{i\in \mathcal{I}}$, timestep $t=0$
\begin{algorithmic}[1]
\WHILE{$t \leq T$}
\STATE TripletCritic receives the memory information $m_t$ and the current state $s_t$
\STATE Generate suggestion for all actors $su = \{su_t^i\}_{i \in \mathcal{I}}$ through Internal Feedback (Algorithm~\ref{alg:IF})
\STATE Genearate the joint action $\mathbf{a}_t=\{a_t^1, a_t^2,\dots,a_t^n\}$ through External Feedback (Algorithm~\ref{alg:EF})
\STATE Execution the joint action $\mathbf{a}_t$, obtain reward $r^i_{t}$ and environmental state $s_{t+1}$
\STATE Collect trajectories $\tau^i$, push transitions $\{(s_t, a^i_t, r^i_t, s_{t+1}\}$ into $\mathcal{M}$
\ENDWHILE
\end{algorithmic}
\end{algorithm}

\begin{algorithm}[H]
\caption{Internal Feedback}
\label{alg:IF}
\textbf{Input}: Maximum number of internal feedback iterations $IF$, current iteration number$f_i=0$, feedback information$F_{if} = None$, state $s_t$, memory $m_t$
\begin{algorithmic}[1]
\WHILE{$f_i \leq IF$}
\FOR{critic $j = 1$ \textbf{to} $2$}
\STATE Generate actions $\mathbf{a}_j = \{a_t^i\}_{i\in\mathcal{I}}$ corresponding to preference $\mathbf{a}_j \sim \mathbf{LLM}_{critic_j}(m_t,s_t, F_{if})$
\ENDFOR
\STATE Assessor makes \emph{Veracity Scrutiny}, get results $vs$
\IF{$vs$ is True}
\STATE Assessor makes \emph{Belief Correction}, generate final action suggestion for all actors $su = \{su_t^i\}_{i \in \mathcal{I}}$, where $su \sim \mathbf{LLM}_{assessor}(m_t,s_t, \mathbf{a}_1, \mathbf{a}_2)$
\STATE $\textbf{break}$
\ELSE
    \STATE Generate feedback information $F_{if}$
\ENDIF
\STATE $f_i = f_i + 1$
\ENDWHILE
\end{algorithmic}
\end{algorithm}

\begin{algorithm}[H]
\caption{External Feedback}
\label{alg:EF}
\textbf{Input}: Maximum number of Enternal feedback iterations $IF$, current iteration number $f_e=0$, feedback information $F_{ef} = []$, suggestion from TripletCritic $su$, observation $\{o_t^i\}_{i\in \mathcal{I}}$, state $s_t$
\begin{algorithmic}[1]
\WHILE{$f_e \leq EF$}
\FOR{agent $i = 1$ \textbf{to} $N$}
\STATE Actor $i$ makes \emph{Plan Confirmation}
\IF{Execute}
\STATE $a_t^i = su_t^i$
\ELSE
\STATE Generate actor feedback Information $F^i_{ef} \sim \mathbf{LLM}_{actor_i}(o^i_t,su^i_t)$
\STATE $F_{ef} = F_{ef} + F^i_{ef}$
\ENDIF
\ENDFOR
\IF{$F_{ef}$ is not $[]$}
\STATE Assessor regenerates action suggestions $su \sim \mathbf{LLM}_{assessor}(m_t,s_t, F_{ef})$
\ELSE
\STATE $\textbf{break}$
\ENDIF
\STATE $f_e = f_e + 1$
\ENDWHILE
\end{algorithmic}
\end{algorithm}

\subsection{Prompt Example in llamac}
\begin{figure*}[htbp]
    \centering
    \includegraphics[width=7 in]{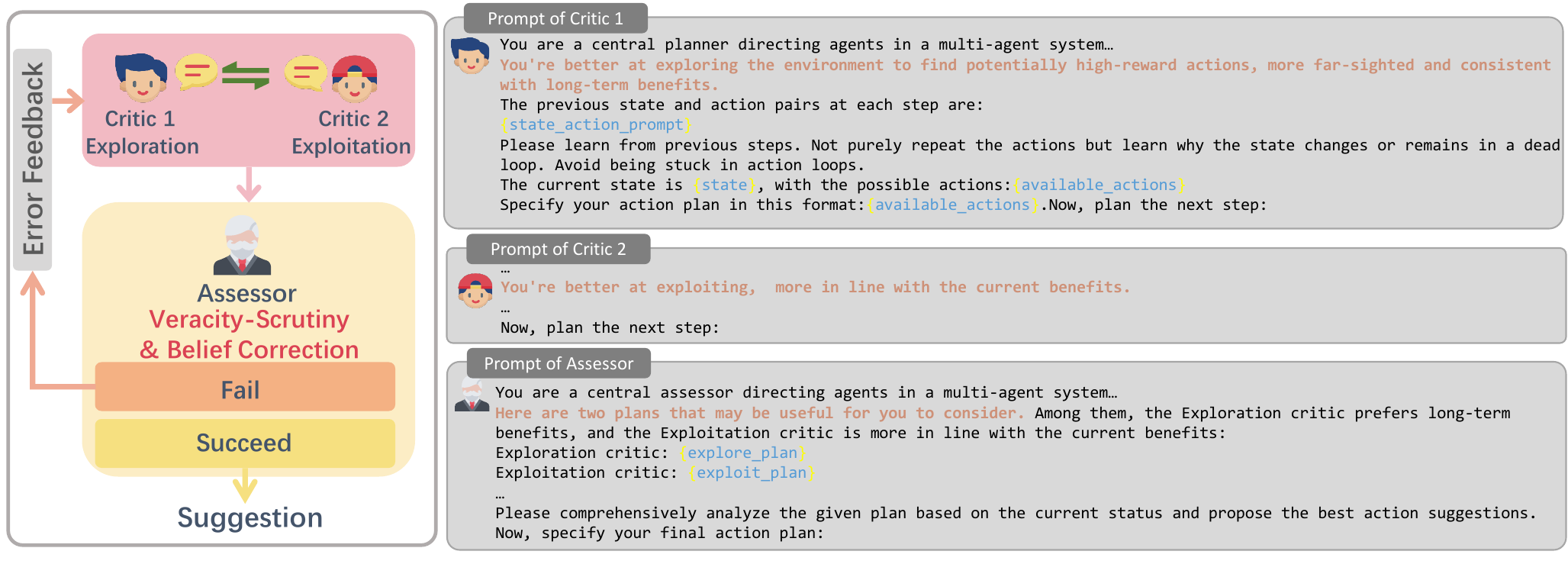}
    \caption{Prompt for the large language model in the internal feedback mechanism.}
    \label{fig:IF_prompt}
\end{figure*}

\begin{figure*}[htbp]
    \centering
    \includegraphics[width=7 in]{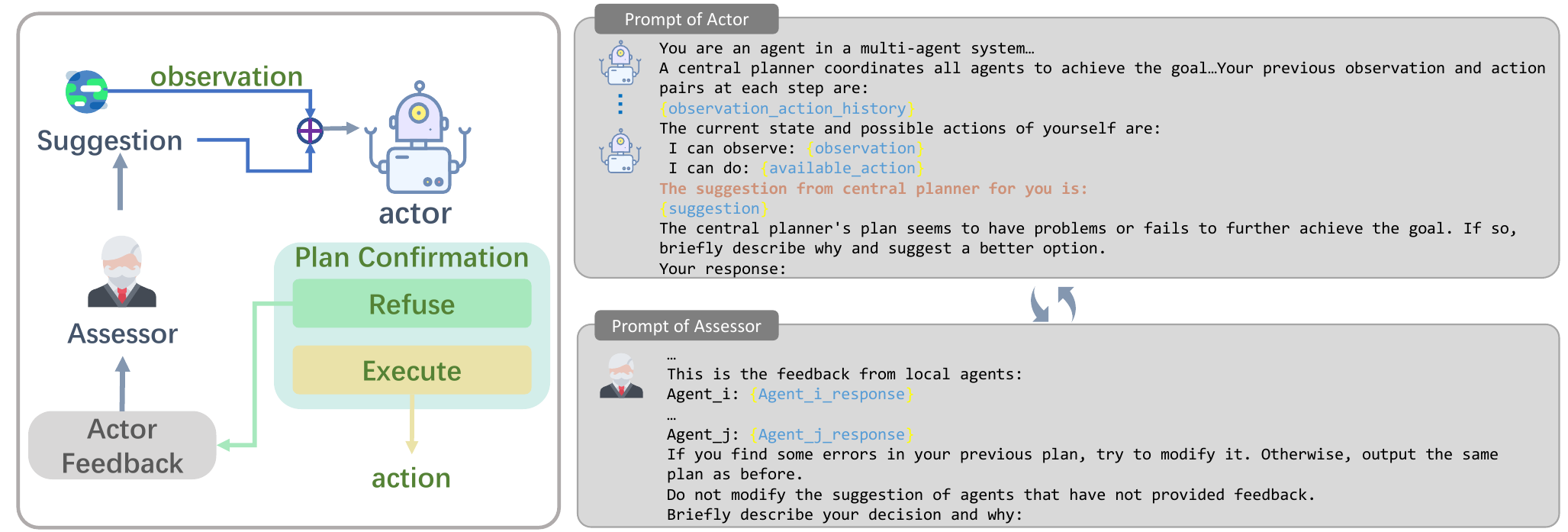}
    \caption{Prompt for the large language model in the external feedback mechanism. Only actors that are determined by the plan confirmation to be reused will execute this access.}
    \label{fig:EF_prompt}
\end{figure*}

As shown in Section~\ref{sec:llamac_pc}, the entire process of LLaMAC's iterative decision-making is facilitated by internal and external feedback mechanisms, enabling seamless collaboration among its modules to accomplish decision tasks in large-scale intelligent agent systems. As shown in Figure~\ref{fig:IF_prompt} and Figure~\ref{fig:EF_prompt}, we demonstrate the core content of the prompt used by LLaMAC to access LLM during the decision-making process.

\

\section{Environment details}
\subsection{System Resource Allocation}
The system resource allocation environment can be regarded as an optimization problem or a single-step decision problem, where the available actions of all agents are fixed at each decision-making instance. 
The memory stores the observation history of the agents in the form of a dictionary: \emph{[\{action:[], system\_reward:[]\}, ..., \{action:[], system\_reward:[]\}]}. Additionally, we require the decision-makers to simultaneously output \emph{thoughts} and \emph{actions} to enhance the reasoning capability of the language model.

\subsection{Grid Transportation}
Grid transportation tasks are inherently more complex and demand higher decision-making capabilities. They involve language models assuming different roles to collaborate through continuous dialogue and interaction, generating long-term action trajectories, and ultimately achieving the final objectives.

\begin{figure*}[htbp]
    \centering
    \includegraphics[width=7 in]{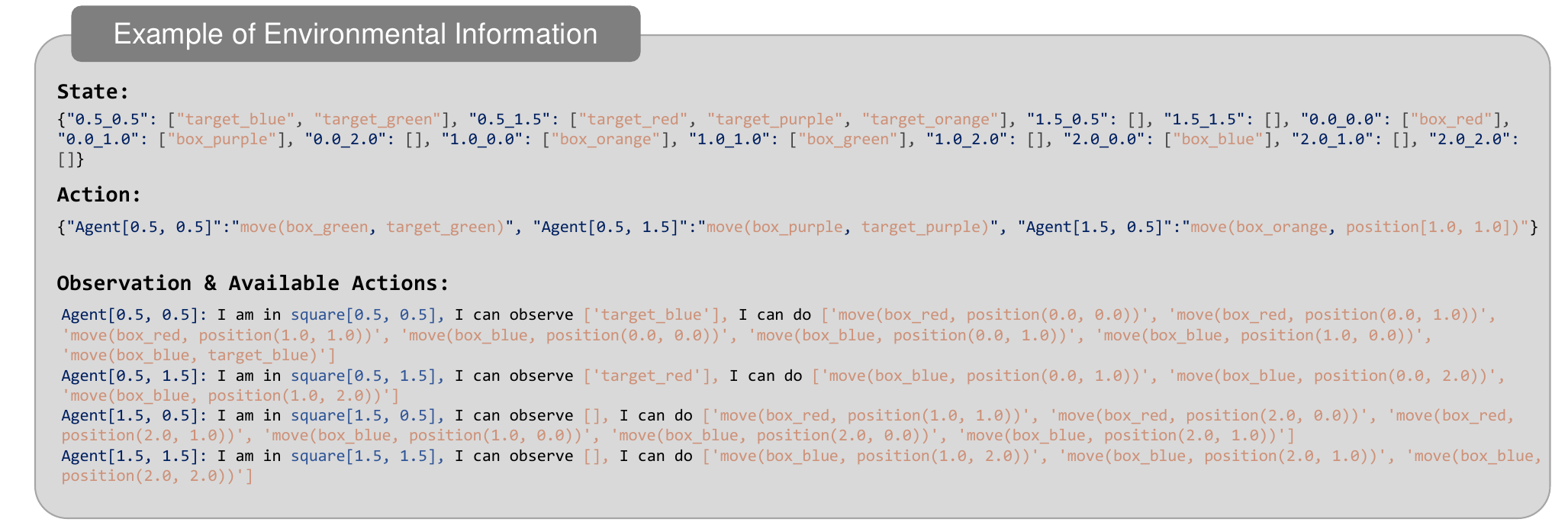}
    \caption{Text description of states, observations, and actions in the grid transport environment.}
    \label{fig:GT_env_prompt}
\end{figure*}

In this environment, the \emph{Veracity Scrutiny} within the Internal Feedback involves policy checks of the joint strategy and is set to evaluate (1) whether the output grammar conforms to the specified format and (2) whether the joint actions result in conflicts. 
The \emph{Plan Confirmation} within the External Feedback involves policy checks specific to each agent and is set to evaluate (1) the availability of actions and (2) whether the suggestions result in a shorter Manhattan distance between objects and targets.
Taking the Hard scenario as an example, the variables utilized in the decision-making process of the intelligent agents are depicted in Figure~\ref{fig:GT_env_prompt}.

\end{document}